\documentclass[letterpaper, 10 pt, conference]{ieeeconf}  

\IEEEoverridecommandlockouts                              

\usepackage{graphics} 
\usepackage{epsfig} 
\usepackage{mathptmx} 
\usepackage{times} 
\usepackage{amsmath} 
\usepackage{amssymb}
\usepackage{bm}
\usepackage{algorithm}
\usepackage{algpseudocode}
\usepackage{multirow}
\usepackage{graphicx}
\usepackage{booktabs}
\usepackage{hyperref}
\hypersetup{
  colorlinks = true, 
  urlcolor = black, 
  linkcolor = black, 
  citecolor = black 
}

\title{\LARGE \bf
An Experimental Modular Instrument With a Haptic Feedback Framework for Robotic Surgery Training*
}

\author{Walid Shaker and Mustafa Suphi Erden
\thanks{*This study was partially funded by the Engineering and Physical Sciences Research Council of the UK, under the EPSRC Grant EP/Y017307/1.}
\thanks{Walid Shaker and Mustafa Suphi Erden are with the School of Engineering and Physical Sciences, Heriot-Watt University, Edinburgh EH14 4AS, UK
{\tt\small (wkhs2000@hw.ac.uk; m.s.erden@hw.ac.uk)}}
\thanks{Multimedia material is available at: \href{https://youtu.be/5SL3UjaDPYU}{https://youtu.be/5SL3UjaDPYU}}
}

\begin{document}
\maketitle
\thispagestyle{empty}
\pagestyle{empty}

\begin{abstract}

Robotic-assisted surgery offers significant clinical advantages but largely eliminates direct haptic feedback, increasing the risk of excessive tool–tissue interaction forces. Although recent commercial systems have begun to introduce force feedback, their high cost limits accessibility, particularly for surgical training. This paper presents a modular experimental robotic laparoscopic instrument integrated with a real-time haptic feedback framework. The proposed instrument employs a wrist-mounted force/torque (F/T) sensor to estimate tool–tissue interaction forces while avoiding the durability and integration challenges of tip-mounted sensors. A haptic feedback framework is developed to extract the external contact forces, render them to the haptic device, and generate stable and perceptually meaningful feedback. The instrument is integrated into the robotic surgery training system (RoboScope) and evaluated through a controlled user study involving a force regulation task. Experimental results demonstrate that haptic feedback significantly improves task success rate, force regulation accuracy, and task efficiency compared to visual-only feedback. The proposed instrument enables stable, high-fidelity haptic interaction, supporting effective robotic surgery training.

\end{abstract}

\section{INTRODUCTION}

Over the past decade, robotic surgery has experienced a substantial increase in adoption, driven by its distinct advantages over both open surgery and conventional laparoscopy \cite{chahal2023transfer}. As a minimally invasive surgical (MIS) procedure, robotic surgery is performed through several small incisions, similar to conventional laparoscopy, and avoids the large incisions required in open procedures. From a clinical standpoint, it preserves the well-established benefits of laparoscopic surgery, including reduced postoperative pain, faster recovery times, shorter hospital stays, and a quicker return to normal activities \cite{lunardi2024robotic}-\cite{kawka2023laparoscopic}. In addition, robotic systems address key limitations of conventional laparoscopy by mitigating hand tremor, eliminating the fulcrum effect, and enhancing motion scaling \cite{wong2024manipulation}. They further provide high-definition three-dimensional visualization, increased instrument dexterity, and improved surgeon ergonomics \cite{lai2024clinical}. 

Despite these advantages, the progression from open surgery to laparoscopic and subsequently to robotic-assisted surgery has been accompanied by a gradual loss of haptic feedback. In open surgery, surgeons maintain direct physical contact with tissues, allowing for a natural sense of touch and perception. This sensory input is significantly diminished in conventional laparoscopy, where elongated instruments constrain direct interaction with the operative field, reducing the surgeon’s ability to perceive tissue texture \cite{colan2022review}. In robotic surgery, this sense of touch is almost entirely absent, as the surgeon operates remotely through robotic interfaces, relying solely on visual cues to assess interaction with tissues \cite{othman2022tactile}-\cite{colan2024tactile}. 

The loss of haptic feedback in robotic-assisted surgery can lead to the application of excessive or traumatic forces, thereby increasing the risk of tissue damage, suture breakage, or surgical errors. To mitigate these risks and enhance patient safety, extensive research has been directed toward developing haptic-integrated interfaces that restore tactile sensation to the surgeon’s console \cite{bergholz2023benefits}.

\section{BACKGROUND AND RELATED LITERATURE}
\label{sec:LR}

The term haptic broadly encompasses both kinesthetic feedback, which is related to the force and motion perception, and cutaneous (tactile) feedback, which involves sensations such as texture, vibration, and skin stretch \cite{patel2022haptic}. Recent innovations include the da Vinci 5 system from Intuitive Surgical, which features force feedback technology that allows surgeons to sense push-pull forces and tissue tension during procedures \cite{rae2025feel}. Additionally, the Senhance® Surgical System from Asensus Surgical is the first of its kind to offer integrated tactile feedback \cite{fujii2025comparison}. 

While these fully fledged robotic systems offer substantial clinical advantages, they require extensive training and the development of a unique set of technical and cognitive skills. The introduction of haptic feedback adds a new layer of complexity that surgeons must master, further elevating the training demands. The high costs associated with acquiring and maintaining these systems present significant barriers for many healthcare institutions. Integrating haptic feedback also requires extensive redesign of both hardware and software, increasing the system’s overall cost. These increased costs can limit access to training opportunities for novice surgeons and restrict the widespread adoption of robotic surgical technologies, especially in resource-limited settings \cite{eckhoff2023costs}.

\begin{figure}[t!]
\vspace{1.5mm}
\centering
\includegraphics[scale=0.35]{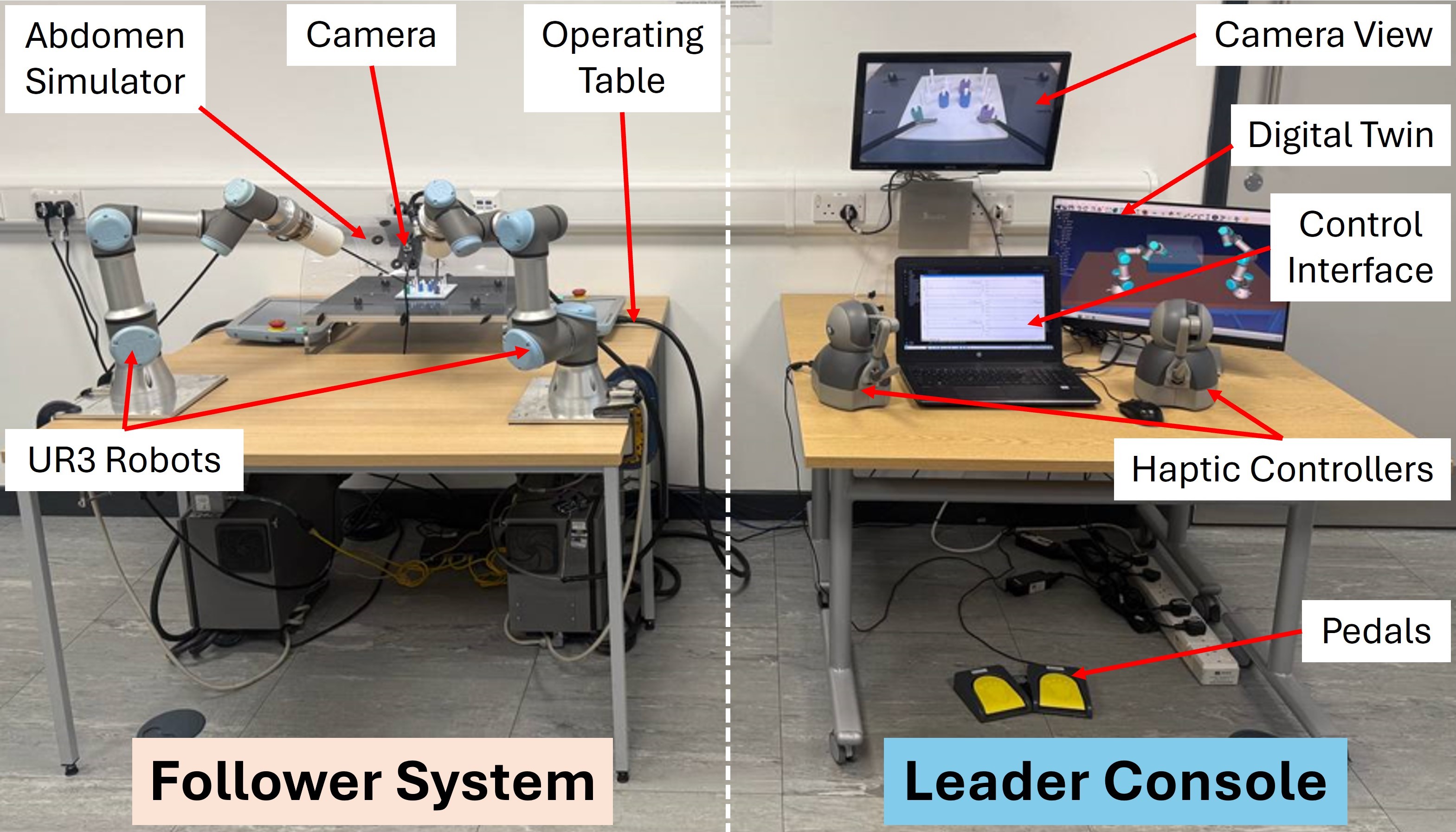}
\caption{Leader and follower sides of the RoboScope system - a low-cost robotic surgery training setup that provides haptic feedback, integrates a digital twin, and supports both on-site and off-site training modalities.}
\label{fig:RoboScope}
\vspace{-1.5mm}
\end{figure}

To address these challenges, we previously introduced RoboScope \cite{shaker2025developing}, shown in Fig.~\ref{fig:RoboScope}, a low-cost robotic surgery training setup designed to promote broad accessibility to robot-assisted surgery training and research. The system supports both on-site and remote teleoperation modes, aiming to reduce barriers to training by providing an affordable alternative that emulates key functionalities of high-end robotic surgical systems. RoboScope enables users to practice a variety of laparoscopic tasks and develop their robotic surgery skills within a realistic and controlled training environment. The system offers key features, including automated performance assessment software, as well as a synchronized digital twin environment that enables real-time visualization and safe operation monitoring.

In this work, we extend the RoboScope system by incorporating haptic feedback, particularly force feedback, to help trainees develop skills aligned with the latest advancements in robot-assisted surgery.

The key contributions of this paper are as follows:
\begin{enumerate}
\item A modular experimental robot-assisted laparoscopic instrument capable of providing real-time haptic sensation in the form of force feedback.
\item A modular haptic feedback framework based on wrist-mounted F/T sensing for robotic surgery training and research.
\end{enumerate}

The remainder of this paper is organized as follows. Section~\ref{sec:Instrument} presents the proposed instrument and its key capabilities. Section~\ref{sec:Haptic} details the haptic feedback framework and its integration with the instrument. Section~\ref{sec:Results} reports the experimental evaluation of the proposed framework through a comprehensive user study. Finally, Section~\ref{sec:Conc} concludes the paper and discusses future work directions.

\section{EXPERIMENTAL SURGICAL INSTRUMENT}
\label{sec:Instrument}

\begin{figure}[t!]
\vspace{1.5mm}
\centering
\includegraphics[scale=0.31]{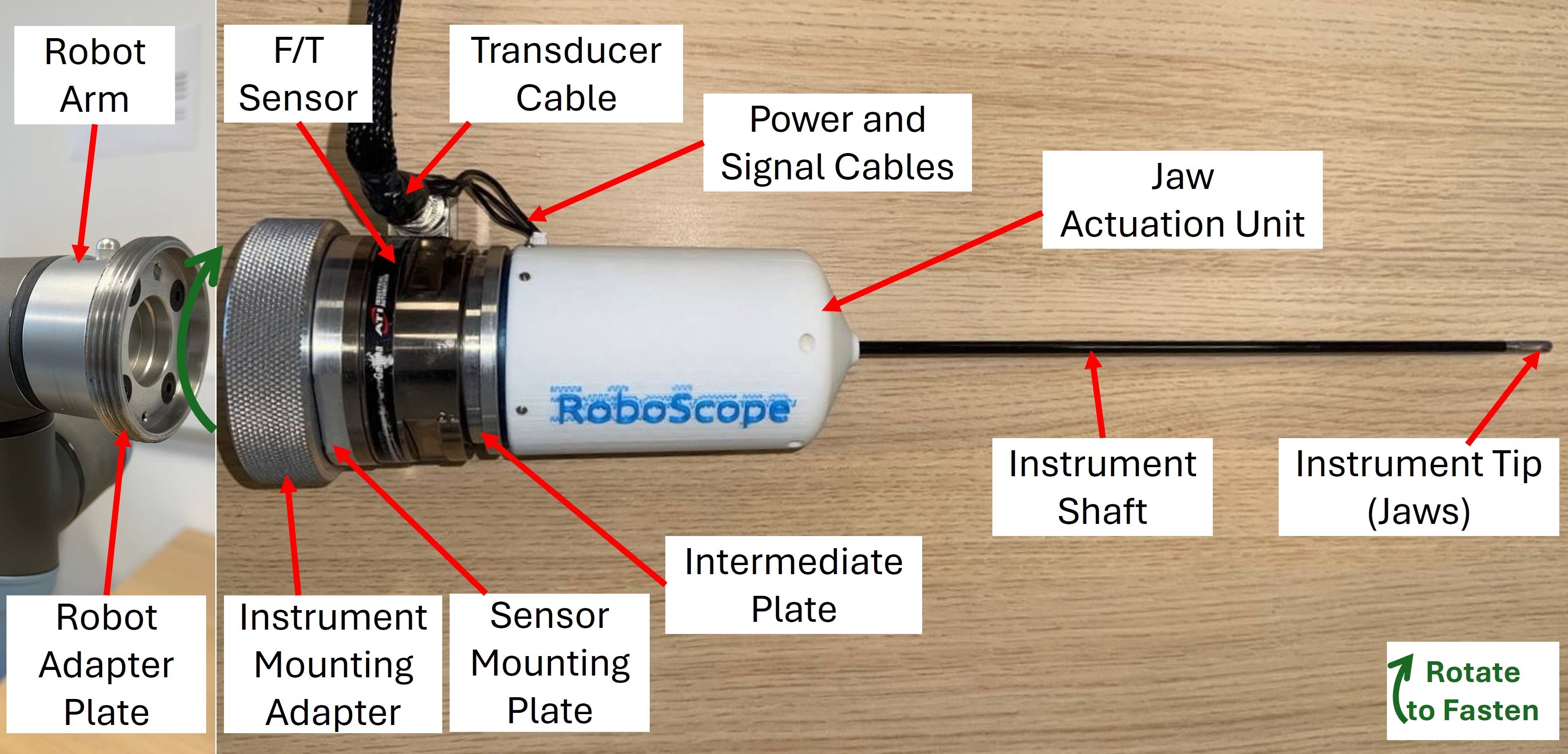}
\caption{Components of the proposed haptic-enabled instrument and its assembly onto the robotic arm, featuring ease of integration and removal and supporting the replacement of the tip with various laparoscopic tools.}
\label{fig:Instrument}
\vspace{-1.5mm}
\end{figure}

The proposed instrument, illustrated in Fig.~\ref{fig:Instrument}, features a robust modular assembly that includes an instrument mounting adapter for secure attachment to the robot, followed by a sensor mounting plate that holds the F/T sensor. This setup is then connected to an intermediate plate that supports the jaw actuation unit, which in turn links to the instrument shaft and tip. A key advantage of this design is its ease of integration and removal from the robot to facilitate efficient maintenance. The instrument can be quickly attached by simply rotating its threaded mounting adapter onto the corresponding robot adapter plate attached to the robot wrist. Once installed, it is crucial to precisely define the Tool Center Point (TCP) at the instrument tip through a calibration of the TCP position and orientation to ensure precise control of the instrument tip relative to the robot base frame.

The key features of the proposed instrument are its actuation and force-sensing capabilities, which are described in detail in the following sections.

\subsection{Actuation Capability}

The core idea behind the design was to transform a traditional laparoscopic instrument into a motorized version suitable for integration into a robotic arm. The development process involved three iterative cycles of design and prototyping, as detailed in \cite{shaker2025developing}. The instrument was designed using CAD software and developed in accordance with Design for Manufacturing (DFM) and Design for Assembly (DFA) principles, with the aim of closely replicating the functionality of fully developed robotic laparoscopic instruments. The instrument features a modular architecture with an interchangeable instrument tip. This allows for easy replacement with various types of laparoscopic tools, such as graspers, scissors, or needle drivers, without compromising its integrated force-sensing capability.

Traditional laparoscopic instruments provide up to five degrees of freedom (DOF), which include insertion (forward and backward), pivoting (pitch and yaw) at the trocar entry point, rotation around the instrument shaft axis, and jaw actuation for opening and closing the instrument tip. All of these movements are manually controlled by the surgeon without any motorization or actuation. In contrast, the proposed motorized instrument allows robotic control of these DOF. The first three, corresponding to insertion and pivoting, are managed by the robot arm joints. The fourth DOF, which is the rotation of the instrument shaft, is independently controlled by the robot’s sixth joint, while the fifth DOF is controlled by a dedicated jaw actuation mechanism \cite{shaker2025developing}.

\begin{figure}[t!]
\vspace{1.5mm}
\centering
\includegraphics[scale=0.42]{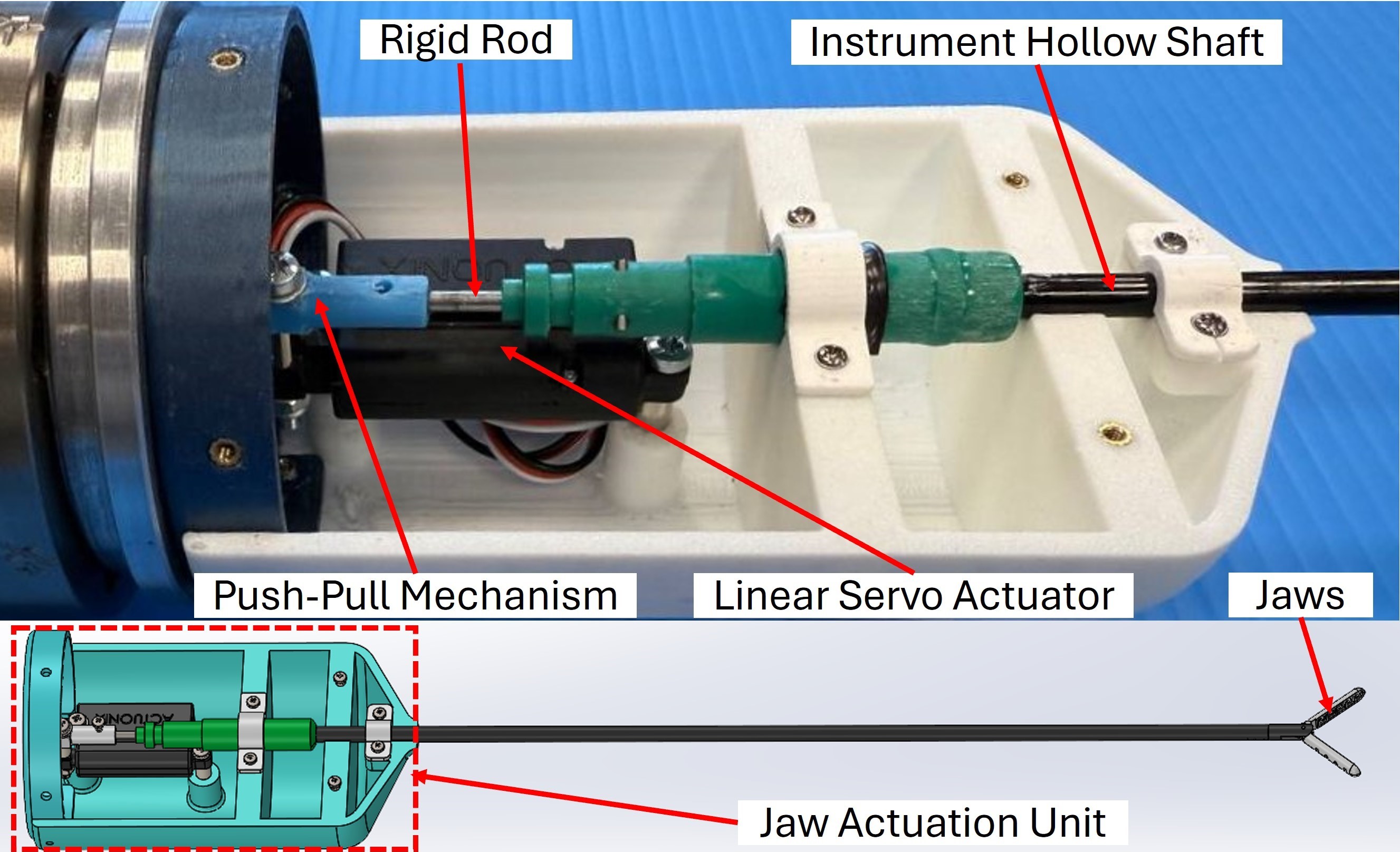}
\caption{Jaw actuation unit showing the rod push-pull mechanism driven by a miniaturized linear servo actuator, which transmits motion through a hollow shaft to enable precise opening and closing of the instrument jaws.}
\label{fig:Unit}
\vspace{-1.5mm}
\end{figure}

The jaw actuation unit employs a miniaturized linear servo actuator, controlled by an Arduino Uno via a serial interface that receives commands to operate the jaws. Opening and closing motions are achieved through a push–pull rod mechanism. The actuator’s linear displacement is transmitted through a rigid rod housed within the instrument hollow shaft, effectively converting axial motion into jaw actuation, as illustrated in Fig.~\ref{fig:Unit}.

For improved usability, the system includes two dedicated buttons, one for fully closing and another for fully opening the jaws. This reduces the likelihood of accidental object release and minimizes repositioning time. The control program also supports variable grip positioning to accommodate different object sizes and protect the actuator from mechanical overstrain. The control logic incorporates both stepwise and continuous movements. A short button press triggers incremental opening or closing by detecting the signal’s rising edge, while a long press (exceeding 0.3 seconds) enables continuous actuation. Positional limits are implemented in the firmware to halt jaw movement when the mechanism reaches its maximum or minimum position.

\subsection{Force-Sensing Capability}

To enable force sensing, the proposed instrument integrates a F/T sensor directly at the robot wrist rather than at the instrument tip. This approach mitigates many of the integration, durability, and maintenance challenges that are common with tip-mounted sensors, particularly for instruments that are disposable or intended for single use. While this wrist-mounted placement simplifies hardware management, it introduces additional complexities. First, the sensor no longer measures forces directly at the tip, where tool-tissue interaction occurs. Second, the raw sensor readings include both the actual external contact forces and internal non-contact contributions, including gravitational effects and sensor mechanical bias. Accurate estimation of external contact forces therefore requires compensating for these effects. In our previous work, we developed a robust real-time algorithm to compensate for these non-contact effects, enabling reliable extraction of the actual external contact forces \cite{shaker2026realtime}. In this work, we build upon that foundation by demonstrating how these compensated forces are utilized to deliver haptic feedback in a robotic surgery training environment, as detailed in the following section.

\section{PROVIDING HAPTIC FEEDBACK}
\label{sec:Haptic}

\begin{figure}[t!]
\vspace{1.5mm}
\centering
\includegraphics[scale=0.43]{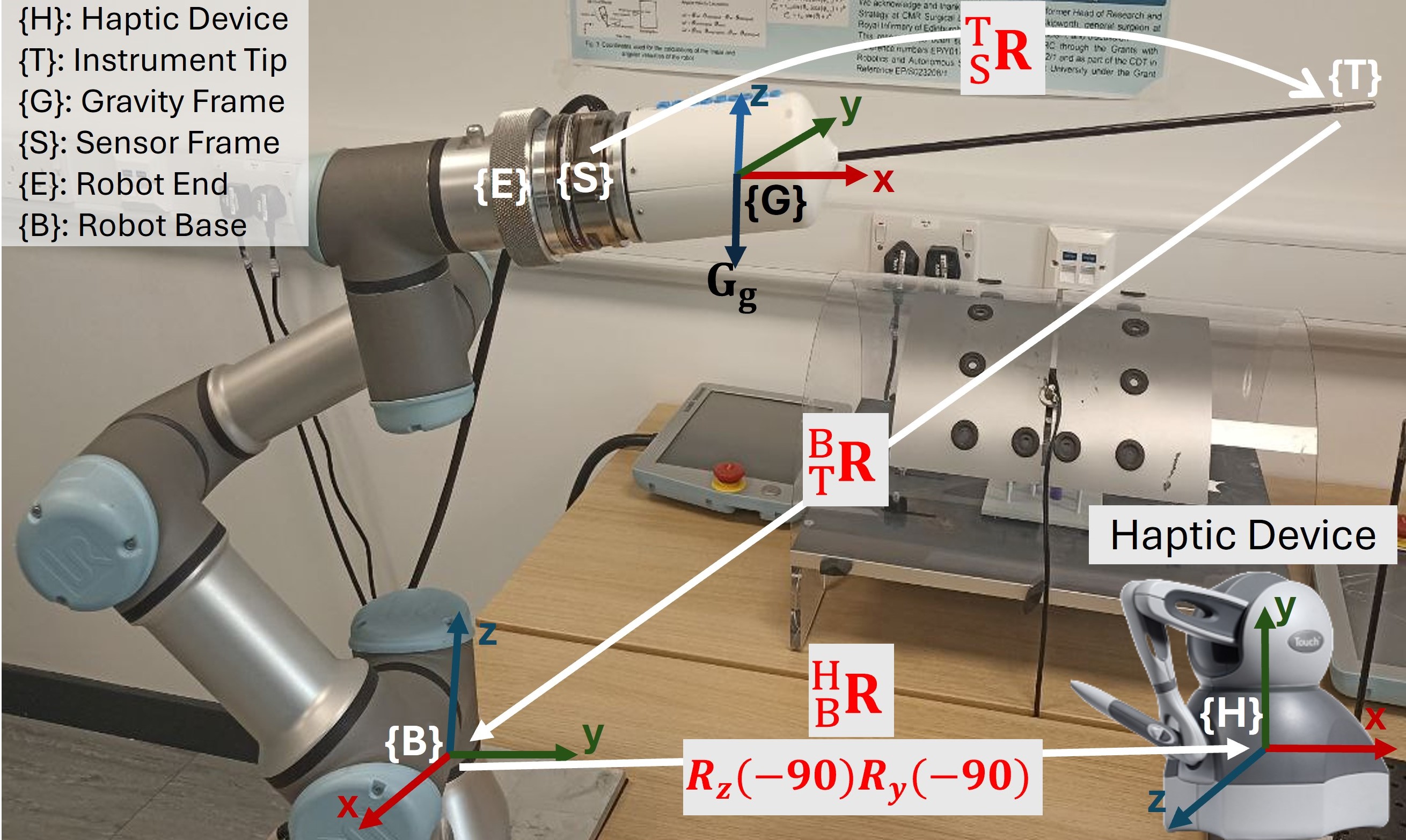}
\caption{The complete mapping between coordinate frames required to transform the contact force from the wrist-mounted sensor to the instrument tip and finally to the haptic device for force feedback provision.}
\label{fig:Mapping}
\vspace{-1.5mm}
\end{figure}

Once gravitational loads and sensor biases are effectively compensated, the isolated true external contact force can be accurately estimated. This compensated force is first smoothed using a moving average filter. It is important to apply filtering after compensation, as directly filtering raw sensor measurements could distort the underlying relationships that the compensation algorithm is designed to capture.

As the force is initially expressed in the sensor frame $\{\mathrm{S}\}$, it must first be used to compute the interaction force at the instrument tip frame $\{\mathrm{T}\}$, where the TCP is located. Then, it is transformed into the robot base frame $\{\mathrm{B}\}$ and subsequently mapped to the haptic device frame $\{\mathrm{H}\}$ to provide real-time haptic feedback. 

Since the sensor frame differs from the instrument tip frame, forces and torques must be transformed accordingly. This general wrench transformation is given by \eqref{eq:ji}, which maps the wrench acting at the origin of a coordinate frame $\{\mathrm{i}\}$ to another frame $\{\mathrm{j}\}$:

\begin{equation}
    \begin{bmatrix}
    {}^\mathrm{j}\mathbf{f} \\
    {}^\mathrm{j}\boldsymbol{\tau} 
    \end{bmatrix}
    =
    \begin{bmatrix}
    {}^\mathrm{j}_\mathrm{i}\mathbf{R} & \mathbf{0}_{3 \times 3} \\
    S({}^\mathrm{j}_\mathrm{i}\mathbf{P}) \: {}^\mathrm{j}_\mathrm{i}\mathbf{R} & {}^\mathrm{j}_{i}\mathbf{R}
    \end{bmatrix}
    \begin{bmatrix}
    {}^\mathrm{i}\mathbf{f} \\
    {}^\mathrm{i}\boldsymbol{\tau} 
    \end{bmatrix}
    \label{eq:ji}
\end{equation}

\noindent Here, ${}^\mathrm{j}\mathbf{f}, {}^\mathrm{j}\boldsymbol{\tau}, {}^\mathrm{i}\mathbf{f}, {}^\mathrm{i}\boldsymbol{\tau}$ represent the forces and torques at the origin of the frames $\{\mathrm{j}\}$ and $\{\mathrm{i}\}$, respectively. ${}^\mathrm{j}_\mathrm{i}\mathbf{R}$ is the rotation matrix from frame $\{\mathrm{j}\}$ to $\{\mathrm{i}\}$, and $S({}^\mathrm{j}_\mathrm{i}\mathbf{P})$ is the skew-symmetric matrix of the position vector $\mathbf{P}$ from frame $\{\mathrm{j}\}$ to $\{\mathrm{i}\}$ expressed in the frame $\{\mathrm{j}\}$. 

Since only force feedback is provided, the actual external contact force measured in the sensor frame (${}^\mathrm{S}\mathbf{f}_{\text{ext}}$) is transformed to the instrument tip frame $\{\mathrm{T}\}$ as:

\begin{equation}
    {}^\mathrm{T}\mathbf{f} = {}^\mathrm{T}_\mathrm{S}\mathbf{R} \, {}^\mathrm{S}\mathbf{f}_{\text{ext}} 
    \label{eq:TCP}
\end{equation}

\noindent where ${}^\mathrm{T}_\mathrm{S}\mathbf{R}$ denotes the rotation offset from the instrument tip frame to the sensor frame, obtained from forward kinematics.

Similarly, the transformation to the robot base frame $\{\mathrm{B}\}$ and subsequently to the haptic device frame $\{\mathrm{H}\}$ is given by

\begin{equation}
    {}^\mathrm{B}\mathbf{f} = {}^\mathrm{B}_\mathrm{T}\mathbf{R} \, {}^\mathrm{T}\mathbf{f} \\
    \label{eq:Base}
\end{equation}
\begin{equation}
    {}^\mathrm{H}\mathbf{f} = {}^\mathrm{H}_\mathrm{B}\mathbf{R} \, {}^\mathrm{B}\mathbf{f}
    \label{eq:Haptic}
\end{equation}
    
\noindent where ${}^\mathrm{B}_\mathrm{T}\mathbf{R}$ is the orientation component of the homogeneous transformation from the robot base frame to the instrument tip frame, obtained from forward kinematics, while ${}^\mathrm{H}_\mathrm{B}\mathbf{R}$ is the rotation from the haptic device frame to the robot base frame. The complete mapping between these coordinate frames is illustrated in Fig.~\ref{fig:Mapping}.

Directly applying the true external contact forces to the haptic device may result in uncomfortable or unrealistic sensations and could exceed the device’s safe operating limits. Moreover, because human perception is more sensitive to relative changes in small forces than in large ones, it is essential to scale the feedback into a safe and perceptually meaningful range.

To address this, a nonlinear scaling based on the hyperbolic tangent function ($\tanh$)
is employed. The $\tanh$ function maps input values to the range $(-1,1)$, producing an 
S-shaped curve. For small forces, the response is approximately linear, enabling high 
sensitivity and precision in delicate interactions. For larger forces, the response 
smoothly saturates, which prevents discomfort, provides a natural-feeling feedback, and protects the hardware.

The maximum force is bounded by $f_{\text{max}}$, set to 3 N to comply with the hardware force limits and prevent user fatigue. The scaling parameter $f_{\text{scale}}$ determines how quickly the saturation occurs. It controls the transition between the linear and saturated regions, providing a balance between sensitivity to small forces and protection against large forces. This value was tuned empirically to 7 N to optimize the dynamic range for the user. The scaled feedback force experienced by the user, ${}^\mathrm{H}\mathbf{f}_{\text{scaled}}$, is then computed as:

\begin{equation}
    {}^\mathrm{H}\mathbf{f}_{\text{scaled}} = 
    \frac{{}^\mathrm{H}\mathbf{f}}{\left\lVert {}^\mathrm{H}\mathbf{f} \right\rVert_2}
    \tanh\!\left(\frac{\left\lVert {}^\mathrm{H}\mathbf{f} \right\rVert_2}{f_{\text{scale}}}\right) {f}_{\text{max}}
    \label{eq:H_scaled}
\end{equation}

\begin{algorithm}[!t]
\caption{Real-Time Haptic Feedback Framework Using a Wrist-Mounted F/T Sensor}
\label{alg:haptic}
\begin{algorithmic}[1]
\Require Compensated external contact force in sensor frame ${}^\mathrm{S}\mathbf{f}_{\text{ext}}$, rotation matrices ${}^\mathrm{T}_\mathrm{S}\mathbf{R}$, ${}^\mathrm{B}_\mathrm{T}\mathbf{R}$, ${}^\mathrm{H}_\mathrm{B}\mathbf{R}$, 
scaling parameters $f_{\text{scale}}$, $f_{\text{max}}$
\Ensure User-perceived haptic feedback ${}^\mathrm{H}\mathbf{f}_{\text{scaled}}$
\While{True}
    \State Acquire compensated force ${}^\mathrm{S}\mathbf{f}_{\text{ext}}$
    \State Apply a moving average filter to ${}^\mathrm{S}\mathbf{f}_{\text{ext}}$
    \Statex \quad\;\:\text{\Comment{Transform force through coordinate frames}}
    \State To tool tip ${}^\mathrm{T}\mathbf{f} \leftarrow {}^\mathrm{T}_\mathrm{S}\mathbf{R} \, {}^\mathrm{S}\mathbf{f}_{\text{ext}}$ \eqref{eq:TCP}
    \State To robot base ${}^\mathrm{B}\mathbf{f} \leftarrow {}^\mathrm{B}_\mathrm{T}\mathbf{R} \, {}^\mathrm{T}\mathbf{f}$ \eqref{eq:Base}
    \State To haptic device ${}^\mathrm{H}\mathbf{f} \leftarrow {}^\mathrm{H}_\mathrm{B}\mathbf{R} \, {}^\mathrm{B}\mathbf{f}$ \eqref{eq:Haptic}
    
   \Statex \quad\;\:\text{\Comment{Apply nonlinear scaling for safe feedback}}
    \If{$\left\lVert {}^\mathrm{H}\mathbf{f} \right\rVert_2 > 0$}
        \State ${}^\mathrm{H}\mathbf{f}_{\text{scaled}} \leftarrow
        \frac{{}^\mathrm{H}\mathbf{f}}{\left\lVert {}^\mathrm{H}\mathbf{f} \right\rVert_2}
        \tanh\!\left(\frac{\left\lVert {}^\mathrm{H}\mathbf{f} \right\rVert_2}{f_{\text{scale}}}\right) {f}_{\text{max}}$ \eqref{eq:H_scaled}
    \Else 
        \State ${}^\mathrm{H}\mathbf{f}_{\text{scaled}} \leftarrow \mathbf{0}$
    \EndIf
    
    \State Send ${}^\mathrm{H}\mathbf{f}_{\text{scaled}}$ to haptic device
\EndWhile

\end{algorithmic}
\end{algorithm}

\noindent where the first term preserves the contact force direction, and the tanh term acts as a magnitude scaling factor that asymptotically approaches ${f}_{\text{max}}$. Algorithm~\ref{alg:haptic} summarizes the complete framework for providing real-time haptic feedback using a wrist-mounted F/T sensor.

\section{EXPERIMENTAL INVESTIGATION}
\label{sec:Results}

This section describes the experimental setup and hardware configuration, followed by a user study conducted to evaluate the proposed haptic feedback framework.

\subsection{Experimental Setup}

\begin{figure}[t!]
\vspace{1.5mm}
\centering
\includegraphics[scale=0.49]{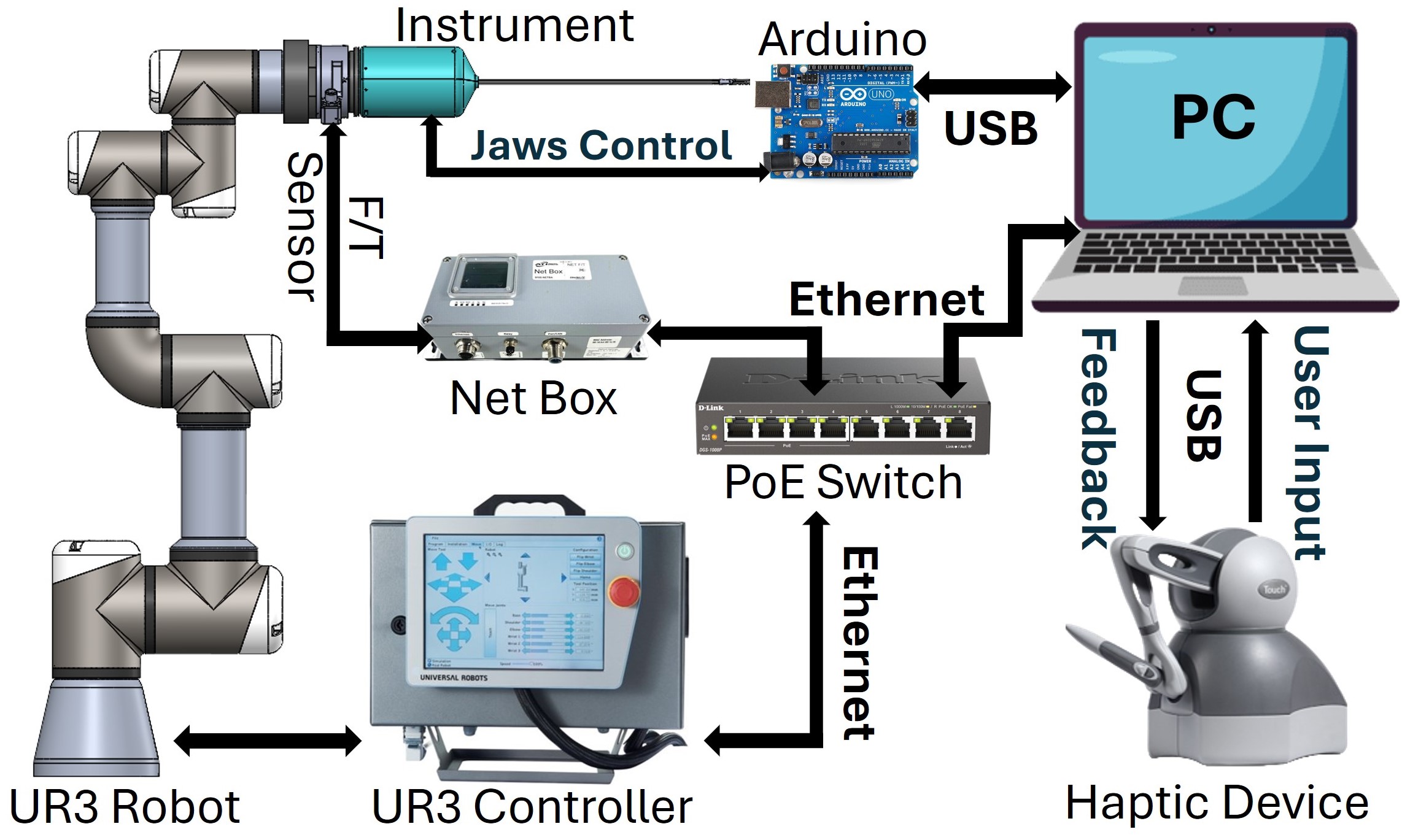}
\caption{RoboScope system architecture showing the proposed instrument with the integrated wrist-mounted F/T sensor and haptic device.}
\label{fig:Setup}
\vspace{-1.5mm}
\end{figure}

The experimental setup for teleoperating the instruments, sensing interaction forces, and delivering haptic feedback is shown in Fig.~\ref{fig:Setup}. Although the RoboScope system includes two robots, each equipped with its own F/T sensor and controlled by a separate haptic device, the system architecture is shown for a single robot for clarity. The same configuration applies to the second robot.

The instrument jaw actuation unit employs the PQ12-63-6-R linear servo actuator from Actuonix Motion Devices, Canada, chosen for its compact size and precise position control. With a maximum no-load speed of 15 mm/s, a maximum lifting force of 45 N, and a 20 mm stroke, it provides reliable and accurate performance for controlling the instrument jaws.

The robots employed in this setup are the UR3 CB3 from Universal Robots, chosen for its safe operation in human–robot collaborative environments. The sensor used is the 6-axis Gamma Force/Torque Sensor (75.4 mm Diameter, 130N / 10Nm Payload) from ATI Industrial Automation, selected for its high strength-to-weight ratio, excellent signal-to-noise performance, and proven suitability for haptic feedback applications. Communication with the ATI sensor is handled using the NetFT Python API, which enables real-time data acquisition via ATI’s Net F/T interface boxes.

Haptic feedback is delivered using the Touch device from 3D Systems. It is a 6 DOF serial kinematic chain, three joints for position control and three for orientation. Only the first three joints are motorized, which is sufficient to provide force feedback that conveys both the magnitude and direction of the interaction forces. Key specifications that informed the selection of the Touch device include its nominal position resolution of 0.055 mm, a maximum exertable force of 3.3 N, and an operating frequency of approximately 1 kHz.

\begin{figure}[t!]
\vspace{1.5mm}
\centering
\includegraphics[scale=0.71]{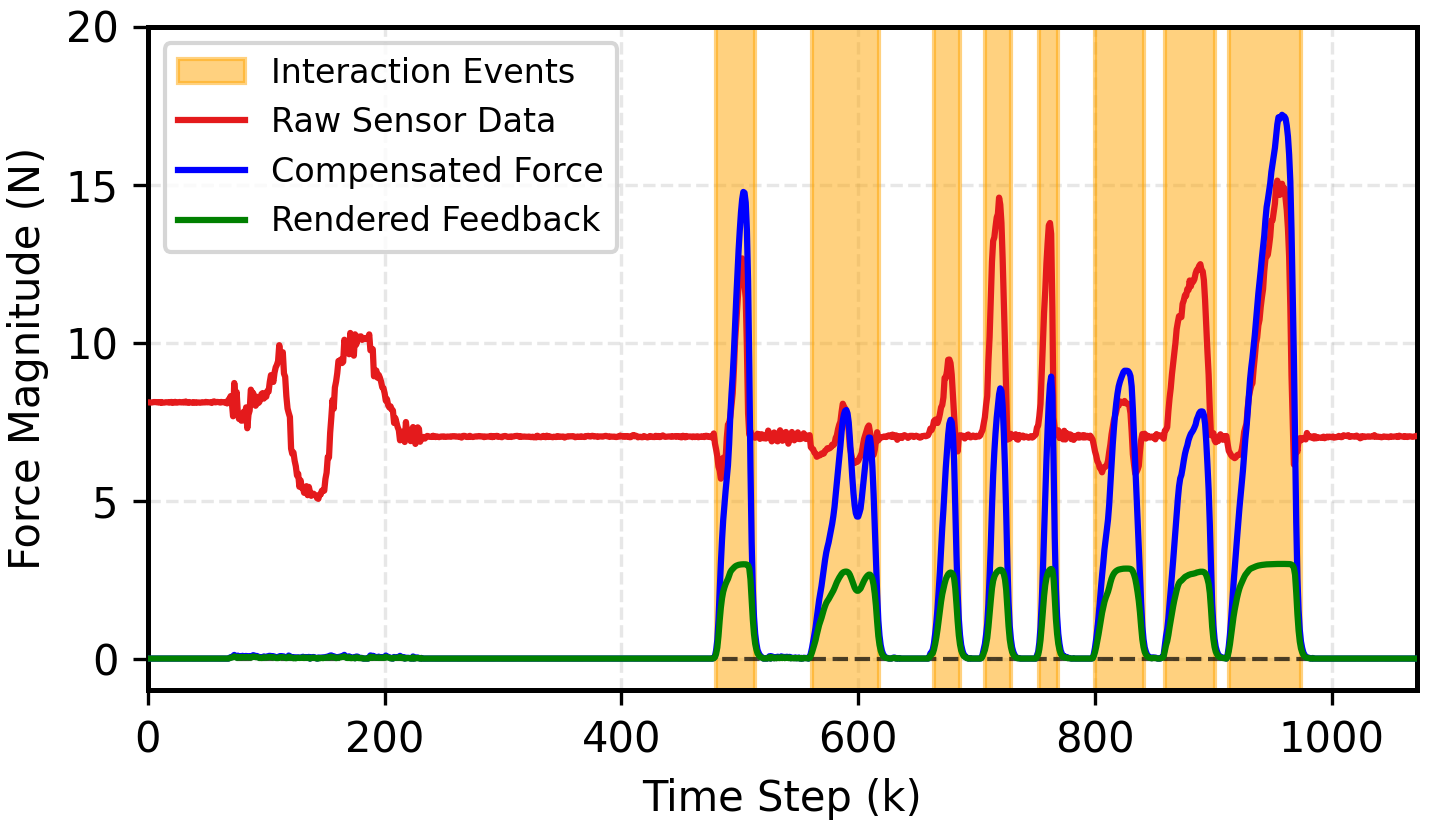}
\caption{Force magnitude analysis across the haptic feedback framework. The plot illustrates the raw sensor data (red), compensated force (blue), and rendered feedback (green), with shaded areas denoting interaction events.}
\label{fig:Robot_R_Magnitude}
\vspace{-1.5mm}
\end{figure}

Utilizing this hardware configuration, the performance of the haptic feedback framework is evaluated by analyzing the resultant force magnitude across the entire sensing-to-feedback pipeline, as illustrated in Fig.~\ref{fig:Robot_R_Magnitude}. The plot shows the raw sensor signal (red), which is corrected for gravitational and bias effects to yield the compensated force (blue). This estimated force is then scaled and transformed to the haptic device as rendered feedback (green). Shaded intervals represent interaction events at the instrument tip, highlighting the framework's ability to isolate and track interaction forces during contact while maintaining a zero-baseline during non-contact periods. The strong correspondence between the estimated and rendered forces indicates that the framework achieves stable, low-latency, and high-fidelity transmission of interaction forces to the user.

\subsection{User Study}

\begin{figure}[t!]
\vspace{1.5mm}
\centering
\includegraphics[scale=0.7]{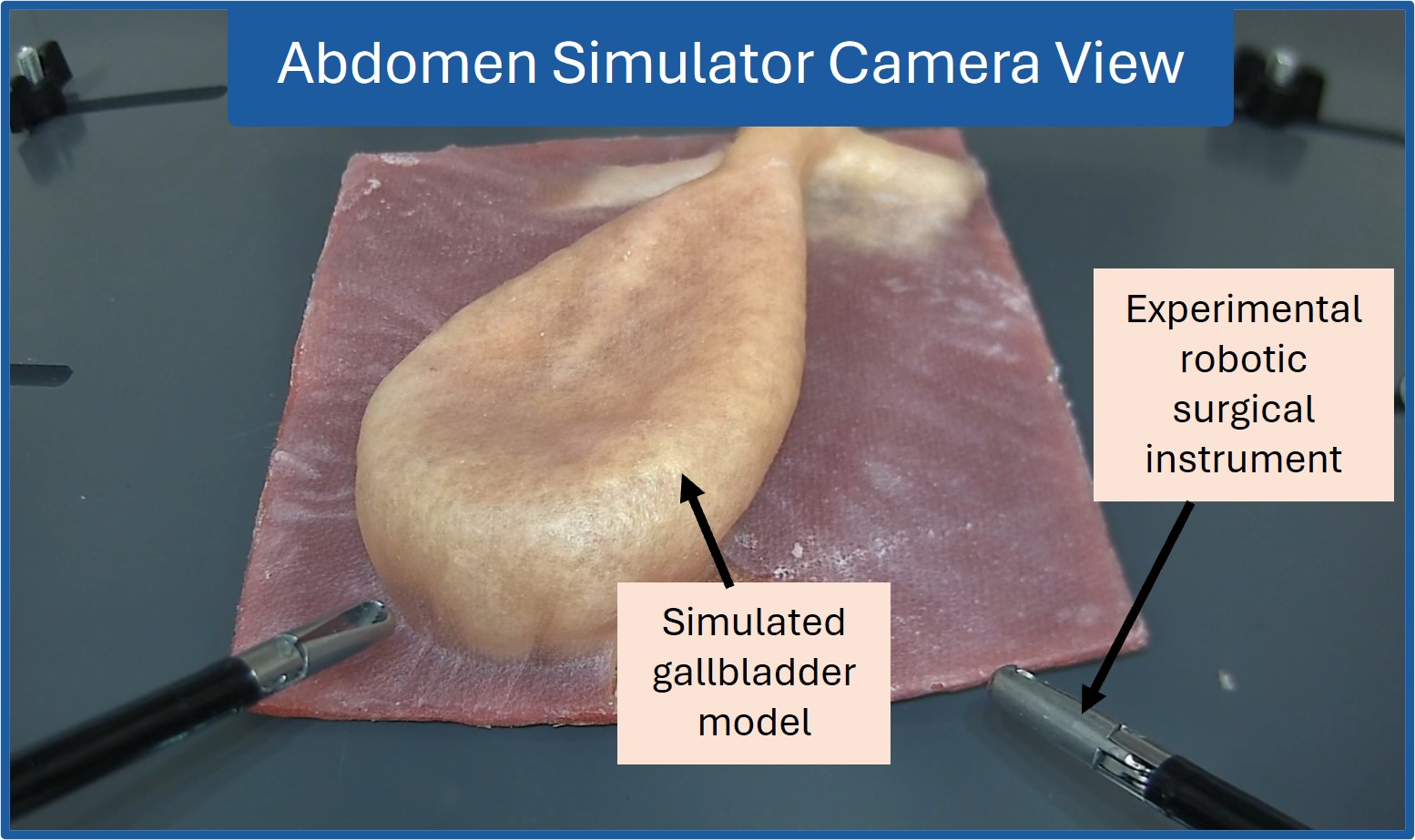}
\caption{The camera view available to the user during the force regulation task, which provides the primary visual cue in both the Haptic OFF and Haptic ON experimental modes.}
\label{fig:Camera_View}
\vspace{-1.5mm}
\end{figure}

To evaluate the effectiveness of the instrument and quantify the impact of haptic feedback on surgical skill acquisition, a user study was conducted with 10 participants. This study was performed following ethics approval (Reference: 2025-13910-18187) granted by the Heriot-Watt University Ethics Committee. The experiment focused on a force regulation task, where the objective was to maintain a low, constant contact force, which is a core task for minimizing tissue trauma in surgery. Participants were required to apply and maintain target force $F_{target}$ for 1 second on a simulated gallbladder model under two experimental modes: 
\begin{itemize}
    \item Haptic OFF: visual cues only from the abdominal simulator camera, shown in Fig.~\ref{fig:Camera_View}.
    \item Haptic ON: Visual cues were provided in addition to haptic feedback.
\end{itemize}

Target forces were qualitatively labeled ‘GENTLE’ (0.6 N) and ‘FIRM’ (1.2 N) to facilitate intuitive interaction. These values were selected within a clinically relevant low-force range to validate the framework under conditions where force discrimination, sensor noise, and contact force estimation accuracy are most critical. Prior to the experiment, users received training in both modes to internalize how these forces felt haptically and appeared visually. The experimental trials were fully automated and randomized, with the computer selecting both the haptic mode and target force. To ensure repeatability, each target force was tested for five trials per mode, resulting in a balanced set of 20 trials per user. 

The target force was displayed only at the start of each trial, and no live force data was shown in either mode. A trial was marked successful when the contact force was held within ±10\% tolerance band around $F_{target}$ for at least 1 second. The trial concluded either when the hold criterion was achieved or the maximum trial time elapsed. Each trial lasted a maximum of 12 seconds; if the target was not achieved within this time, the trial was logged as a failure. A pause of 2 seconds was inserted between trials, and all data logging was performed automatically.

The performance of both modes was evaluated using three objective metrics:
\begin{itemize}
    \item Task Completion Time (TCT): Measures efficiency, defined as the time taken to reach the target and meet the holding criterion.
    \item Root Mean Square Error (RMSE): Measures the overall accuracy and stability of the force throughout the trial.
    \item Maximum Absolute Error (Max AE): Measures the maximum absolute deviation from the target during the trial, representing the peak force excursion.
\end{itemize}

\begin{figure}[t!]
\centering
\includegraphics[scale=0.7]{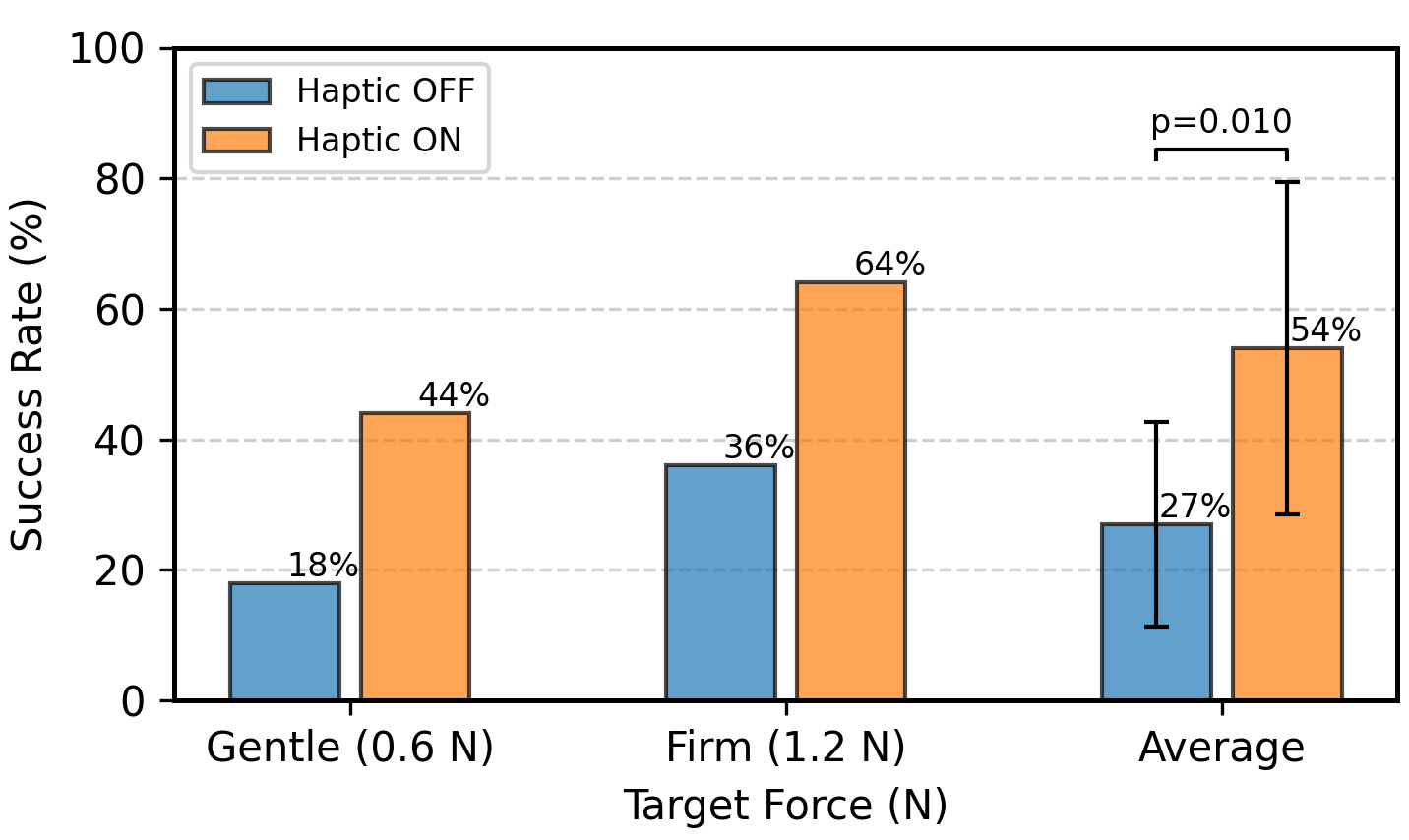}
\caption{Success rate for each target force (GENTLE 0.6 N, FIRM 1.2 N) along with the overall average, representing the percentage of successful trials for both the Haptic OFF and Haptic ON modes across all participants.}
\label{fig:Success_Rate}
\end{figure}

\begin{figure}[t!]
\centering
\includegraphics[scale=0.7]{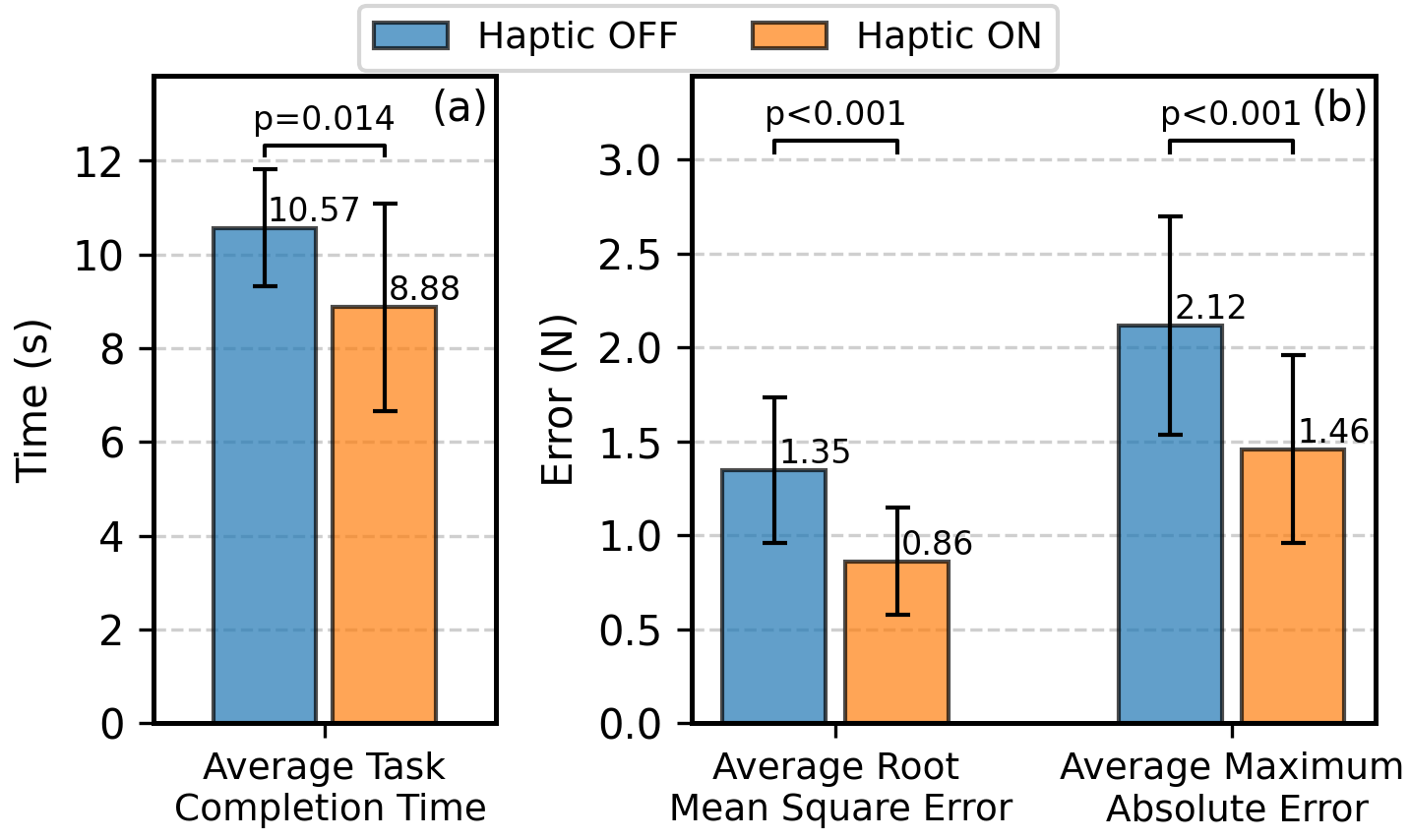}
\caption{Average performance metrics (Haptic OFF vs. ON) across all targets and trials for all participants. (a) Average Task Completion Time. (b) Average Root Mean Square Error and Average Maximum Absolute Error.}
\label{fig:Overall_Metrics}
\end{figure}

As illustrated in Fig.~\ref{fig:Success_Rate}, the average success rate across all target forces doubled when haptic feedback was engaged, rising significantly from 27\% (Haptic OFF) to 54\% (Haptic ON), with standard deviation bars confirming consistency across participants. This substantial improvement demonstrates the effectiveness of the haptic cues in achieving and maintaining the desired contact force. Task efficiency also improved, as shown in Fig.~\ref{fig:Overall_Metrics}(a), where a 16\% reduction in the average TCT was observed, indicating that haptic feedback enabled users to converge on the target force more smoothly with fewer corrective movements.

Force regulation accuracy improved significantly as well, as shown in Fig.~\ref{fig:Overall_Metrics}(b). The average RMSE decreased from 1.35 N (Haptic OFF) to 0.86 N (Haptic ON), corresponding to a 36\% reduction in overall tracking error. This outcome confirms that the haptic cues allowed users to regulate force more accurately and precisely, minimizing unwanted fluctuations. Similarly, the average Max AE reduced from 2.12 N (Haptic OFF) to 1.46 N (Haptic ON), reflecting a 31\% reduction in peak force excursions, which directly relates to safer tissue interaction. A Wilcoxon signed-rank test confirmed significant improvements across all metrics, as indicated in Figs.~\ref{fig:Success_Rate} and \ref{fig:Overall_Metrics}.

In summary, the results show that haptic feedback consistently improved force regulation accuracy, task efficiency, and success rate compared to the visual-only feedback. These initial results confirm that the proposed instrument, integrated with the haptic feedback framework, provides more precise and stable force control, particularly in low-force surgical conditions where tactile sensitivity is critical. By operating at a 1 kHz refresh rate, the proposed system delivers high-fidelity haptic feedback to improve user performance and training quality in robotic surgery environments.

\section{CONCLUSION}
\label{sec:Conc}

This paper presents a modular experimental robotic laparoscopic instrument, supported by a real-time haptic feedback framework designed for robotic surgery training and research. By leveraging a wrist-mounted F/T sensor and a robust compensation strategy, the proposed approach enables accurate estimation of tool–tissue interaction forces without the complexity associated with tip-mounted sensing. The estimated forces are transformed across coordinate frames and rendered to a haptic device using nonlinear scaling, ensuring stable, low-latency, and perceptually meaningful feedback. Experimental evaluation through a controlled user study demonstrated that the inclusion of haptic feedback substantially improves force regulation performance. Participants achieved higher success rates, reduced task completion times, and significantly lower force tracking errors when haptic feedback was enabled, compared to visual cues alone. These results confirm the effectiveness of the proposed framework, showing that an affordable wrist-mounted sensing instrument can deliver reliable force feedback. Future work will focus on expanding the range of training tasks and conducting user studies with surgeons to assess system performance in more complex surgical scenarios.

\addtolength{\textheight}{-12cm}  


\bibliographystyle{IEEEtran}
\bibliography{ref}
\end{document}